\documentclass[10pt, a4paper]{article}
\usepackage{lrec2022} % this is the new LREC2022 Style
\usepackage{multibib}
\newcites{languageresource}{Language Resources}
\usepackage{graphicx}
\usepackage{tabularx}
\usepackage{soul}
\usepackage{xcolor}
\usepackage{flushend}
\usepackage{titlesec}
\titleformat{\section}{\normalfont\large\bfseries\center}{\thesection.}{1em}{}
\titleformat{\subsection}{\normalfont\SmallTitleFont\bfseries\raggedright}{\thesubsection.}{1em}{}
\titleformat{\subsubsection}{\normalfont\normalsize\bfseries\raggedright}{\thesubsubsection.}{1em}{}
\renewcommand\thesection{\arabic{section}}
\renewcommand\thesubsection{\thesection.\arabic{subsection}}
\renewcommand\thesubsubsection{\thesubsection.\arabic{subsubsection}}
\usepackage{epstopdf}
\usepackage[utf8]{inputenc}

\usepackage{hyperref}
\usepackage{xstring}

\usepackage{color}

\usepackage{booktabs}
\usepackage{caption} 
\usepackage{amssymb,amsmath,bm}

\title{Evaluating Gender Bias in Speech Translation}
\name{Marta R. Costa-jussà$^1$, Christine Basta$^{1,2}$, Gerard I. Gállego$^1$}
\address{
    $^1$ TALP Research Center, Universitat Politècnica de Catalunya, Barcelona, Spain \\
    $^2$ Institute of Graduate Studies and Research, Alexandria University, Egypt\\
     \{marta.ruiz, christine.raouf.saad.basta, gerard.ion.gallego\}@upc.edu
}
\abstract{
 The scientific community is increasingly aware of the necessity to embrace pluralism and consistently represent major and minor social groups. Currently, there are no standard evaluation techniques for different types of biases. Accordingly, there is an urgent need to provide evaluation sets and protocols to measure existing biases in our automatic systems. Evaluating the biases should be an essential step towards mitigating them in the systems. This paper introduces WinoST, a new freely available challenge set for evaluating gender bias in speech translation. WinoST is the speech version of WinoMT, an MT challenge set, and both follow an evaluation protocol to measure gender accuracy. Using an S-Transformer end-to-end speech translation system, we report the gender bias evaluation on four language pairs, and we reveal the inaccuracies in translations generating gender-stereotyped translations.
 \\ \newline \Keywords{challenge set, multilinguality, direct speech-to-text translation} }

\begin{document}

\maketitleabstract

\section{Introduction}

There is a massive lack of representation of diverse gender, race, cultural groups in the power structures. These problems permeate data science causing unbalanced progress in this area, which misrepresents social groups by gender, race, and nationality.

%Biases have been shown in machine translation system when translating from high-resource languages compared to low-resourced \cite{aharoni-etal-2019-massively}, and from high-morphological languages compared to low-morpholigal ones producing gender-stereotyped forms in inaccurate translations  \cite{gabriel:2019,prates,saunders_domain_GB:2020}.

Biases have been shown in Machine Translation (MT) systems when translating from high-resource languages to low-resourced ones \cite{aharoni-etal-2019-massively} and from less-morphological languages to high-morphological ones producing gender-stereotyped forms in inaccurate translations \cite{gabriel:2019,prates,saunders_domain_GB:2020}. Additionally, Automatic Speech Recognition (ASR) has demonstrated biases having a higher error rate for female voices than for males \cite{tatman-2017-gender}. Speech Translation (ST) is at the intersection of ASR and MT tasks, perpetuating biases from both tasks.

%Consequently, machine translation systems offer high-quality translations for high-resource languages compared to low-resourced \cite{aharoni-etal-2019-massively}; image recognition systems perform much better for European and American faces \cite{buolamwini}, and automatic speech recognition has a low error rate for female's voices than for male's ones \cite{tatman-2017-gender}. 

While this is still an uncovered challenge in most artificial intelligence tasks, the scientific community is making huge efforts towards bringing to light this challenge. In the short term, researchers are devoting efforts towards bias evaluation, detection, and mitigation methods. On the other hand, there is multidisciplinary work in education, politics, and communications in the long-term perspective.

 %{ \color{red}Speech Translation (ST) is at the intersection of Automatic Speech Recognition (ASR) and Machine Translation (MT) tasks. Therefore, biases can propagate from both tasks; ASR and MT biased outputs  \cite{tatman-2017-gender,Koenecke7684,gabriel:2019,prates,saunders_domain_GB:2020}.}

%Within these tasks, biases have been detected and studied from different perspectives. { \color{orange}For the ASR task, it has been shown that the system tends to produce different outputs with gender and dialects \cite{tatman-2017-gender} and race \cite{Koenecke7684}.On the other hand, as demonstrated by previous studies \cite{gabriel:2019,prates,saunders_domain_GB:2020}, MT can perpetuate gender bias by producing gender-stereotyped forms in inaccurate translations.}

For addressing this problem in MT, some papers propose evaluating gender bias in MT \cite{gabriel:2019,levy_BUG_GB:2021}, creating multilingual balanced data to train fairer systems \cite{costajussa:2020}, or using equalizing techniques to mitigate the effect of unbalanced training data \cite{Font:2019,saunders_domain_GB:2020}. Finally, in ST, \cite{bentivogli:2020_mustshe} have proposed the MuST-SHE corpus, a natural benchmark for automatically evaluating gender bias.

The main contribution of our work is providing a large-scale multilingual ST challenge set,\footnote{Freely available in Zenodo (10.5281/zenodo.4139080)} which follows an evaluation protocol for the analysis of gender bias, previously proposed for MT \cite{gabriel:2019} (see Table \ref{tab:mustshe_winost}). This set mainly evaluates the gender bias by analysing the inflection of the gendered words in the sentences, from English to any other language. Additional bias can be detected considering different speakers' gender, but such bias is out of the scope of our challenge set. Although the synthetic challenge set can propagate stereotypes as described by \cite{blodgett-etal-2020-language}, it still remains valuable to evaluate and monitor the model's performance and provide insights into how the model treats gender-related issues \cite{blodgett_datasets_acl:2021}.
%The benchmark evaluates two categories of gender bias: one related to the speaker gender and another to the utterance content. The benchmark is limited to English-to-French, English-to-Spanish and English-to-Italian language pairs. Differently, 

\begin{table*}[th]
\caption{ST Gender Bias (GB) benchmarks comparison, MuST-SHE and WinoST with respect to the GB in speaker's voice, GB in sentence content and language pairs covered by each benchmark. X stands for any language, meaning that WinoST can be evaluated from English to any other language.}
  \label{tab:mustshe_winost}
  \centering
  \begin{tabular}{lcc}
    \toprule
     & \textbf{MuST-SHE} & \textbf{WinoST} (Ours) \\
   
    \midrule
    \textit{GB in sent. content} & \checkmark & \checkmark \\
    \textit{GB in speaker voice} & \checkmark &  \\
    \midrule
    \textit{Language pairs} & En $\xrightarrow{}$ Fr, Es, It & En $\xrightarrow{}$ X \\
    \bottomrule
  \end{tabular}
    
\end{table*}

\section{Bias Statement}

As proposed in previous work \cite{blodgett-etal-2020-language} and suggested in related venues,\footnote{https://genderbiasnlp.talp.cat/gebnlp2020/how-to-write-a-bias-statement/} we formulate the bias statement of our work. Our work focuses on proposing a challenge set and benefit from an evaluation protocol for the application of ST. Our challenge set serves to measure how accurately gendered source spoken words are translated. It includes stereotypical and anti-stereotypical examples, adding value to the evaluation analysis.

The main contribution of our proposal is to have an objective evaluation protocol that can measure how biased our ST systems are. We consider that our work is based on a synthetic dataset simplified to a binary gender, which does not aim to be a mirror of reality. However, we are contributing to exhibiting that current systems contain biases while sustaining that long-term policies are required to solve this problem \cite{datafeminism}.

\section{Gender Bias within MT systems and Related Work}
\label{sec:relwork}
%line of correcting the gender
%line of detecting issues in translation
%line of challenge sets speech
%why there is gender bias problem
%winomt evaluation
 Grammatical gendered languages have richer grammar for expressing gender. In these languages, gender has to be assigned to all nouns, and consequently, all the articles, verbs, and adjectives have to correspond with the gender of this noun. This leads to incorrect translation from a natural gendered language (English) to a grammatical gendered marking language (Spanish) due to the lack of explicit evidence of the gender in the source \cite{gonen:2020_automatically}. 
Thus, gender bias occurs due to the gender information loss and over-prevalence of gendered forms in the training data. An illustrative example would be the translation from \textit{The doctor spoke to the responsible nurse in the hospital.} to its Spanish version. The Spanish translation would be \textit{El médico habló con la enfermera responsable en el hospital.}, assuming the nurse to be female and the doctor to be male. Although nothing in the text mentions their genders, the systems would translate in a biased manner. Generally speaking, MT systems have proven to have biased outputs. Even in more context, translations seem to ignore the context in sentences with more favoring to the masculine and stereotyped gendered roles \cite{gabriel:2019,saunders_domain_GB:2020}. The main reason for such bias is training models on human-biased data \cite{costajussa:2019}. 

Researchers have recently dedicated efforts to attempt resolving such bias. We can thereby describe three research lines of gender bias in MT: mitigating gender bias in MT, detecting issues in translation, and creating the challenge test sets to evaluate gender bias in the systems.

Regarding the first research line, approaches have been dedicated to solving the translation bias from a neutral gendered language to a grammatically gendered marking one. Adding a gender tag of the speaker during training enhances the translation quality, as demonstrated in \cite{eva:2018}. It facilitates the gender prediction correctly when translating from English to other gendered languages, giving control over the translation hypothesis gender. This was confirmed in recent work by \cite{basta:2020}, who proved that adding gender helps to increase the accuracy of gendered translations. Moreover, the authors showed that increasing context has a better effect on gendered translations, leading to higher performance. \cite{moryossef:2019_fillinggender} incorporate gender information by prepending a short phrase for each sentence in inference time, which acts as a gender label. Recent work \cite{saunders-byrne-2020-reducing,costajussa_adrian_finetuning:2020} has treated gender biasing as a domain-adaptation problem, in which the system is fine-tuned instead of retrained for mitigating gender bias. They have adapted a set of synthetic sentences with equal numbers of entities using masculine and feminine inflections to fine-tune the MT system. \cite{Font:2019} introduced the idea of adjusting the word embeddings, which improved performance on an English-Spanish occupations task.

%these works are useful for speech corpuses

Concerning research for detecting gender translation issues, \cite{prates} have examined Google translations and proved that mentions of stereotyped professions are more reliably translated than those anti-stereotyped. In their study, they have used sentence templates filled with word lists of professions and adjectives. The authors in \cite{cho:2019_measuringGB} also have studied the pronouns' translations for the Korean language using sentence templates. 
The  recent work \cite{gonen:2020_automatically} has proposed a BERT-based perpetuation method to identify gender issues in MT automatically. The technique discovers new sources of bias beyond the word lists used previously.

The most related line of research to our work is creating challenge sets. The first MT challenge set was introduced by \cite{gabriel:2019}, called WinoMT. WinoMT is a test set of 3,888 sentences, where each sentence contains two human entities, where one of them  is co-referent to a pronoun. The evaluation depends on comparing the translated entity with the golden gender,  with the objective of a correctly gendered translation. The authors identified three metrics for the evaluation:  accuracy, \bm{$\bigtriangleup G$} and \bm{$\bigtriangleup S$}. The accuracy is the percentage of correctly gendered translated entities compared to the gender of golden entities. \bm{$\bigtriangleup G$} is the difference in F1 score between the set of sentences with male entities and female entities set. \bm{$\bigtriangleup S$} is the difference in accuracy between the set of sentences with pro-stereotypical entities and the set with anti-stereotypical entities. A pro-stereotypical set identifies `developer' as male and `hairdresser' as female. An anti-stereotypical identifies the former as female and the latter as male. 
As far as we know, the only existing related work for studying gender bias in speech is presented in \cite{bentivogli:2020_mustshe}.
They created a benchmark dataset that 
%of three language pairs (English-Italian/English-French/English-Spanish) and accordingly evaluated their systems. The benchmark 
evaluates two categories of gender bias: one related to the speaker gender and another to the utterance content. The benchmark is limited to English-to-French, English-to-Spanish and English-to-Italian language pairs. 

Given that the closest work to ours is MuST-SHE \cite{bentivogli:2020_mustshe}, as follows we detail the key difference between the two resources: while MuST-SHE contains naturally occurring gender phenomena, WinoST is a synthetic challenge set. This difference has several implications, not only in terms of size of the resources themselves (generating synthetic data is somehow easier than collecting them in the wild) but also in terms of their applicability in realistic evaluation settings. Both types of resources are useful and much needed. Therefore, we see some complementarity here rather than concurrent alternatives where one is better than the other.

\section{Speech Translation System}
\label{sec:speechsystem}
%illustrate the system (Gerard)

We trained a ST system to evaluate its gender bias with the methodology we are presenting. We used an end-to-end ST approach that directly translates the utterance without obtaining the intermediate transcriptions. This task was introduced by \cite{berard:hal-01408086}, and recently it had a growing interest in the research community \cite{Weiss2017,CrossVila2018,Liu2019}.
%,salesky-black-2020-phone}. 
The data we used to train it is the MuST-C corpus, that consists of speech fragments from TED Talks, its transcriptions and translations into 8 European languages \cite{mustc19}.

The architecture we used is the S-Transformer, a popular adaptation of the Transformer for ST \cite{Gangi2019}. It applies a stack of convolutions and self-attention layers to process the log-Mel spectrograms extracted from the speech utterances. The two bidimensional (2D) convolutional layers are in charge of capturing local patterns in the spectrogram, in both time and frequency dimensions. Moreover, they reduce the features maps by four, which is crucial to avoid memory issues that happen when feeding the Transformer with long sequences. Then, the two 2D self-attention layers, which were introduced by \cite{8462506}, model long-range dependencies that convolutional layers cannot capture. Finally, the self-attention layers of the Transformer encoder also include a logarithmic distance penalty that biases them towards the local context \cite{Sperber2018}.

Following a common approach, we pre-trained the S-Transformer encoder for ASR to improve the performance of the final ST system, as introduced by \cite{berard:hal-01709586} and recommended by the authors of the S-Transformer.

\section{Proposed Gender Evaluation: WinoST challenge set}
\label{sec:challengeset}

WinoST is the speech version of WinoMT, recorded in off-voice by an American female speaker, and consists of $3,888$ speech audios in English. By nature, sentences from WinoST contain information in the utterance content, not in gender information in the speaker's voice. An example of these sentences is \textit{The developer argued with the designer because she did not like the design.}, where \textit{she} refers to \textit{developer}, meaning that the \textit{developer} is actually a female.

\begin{figure}[t]
  \centering
  \includegraphics[width=\linewidth]{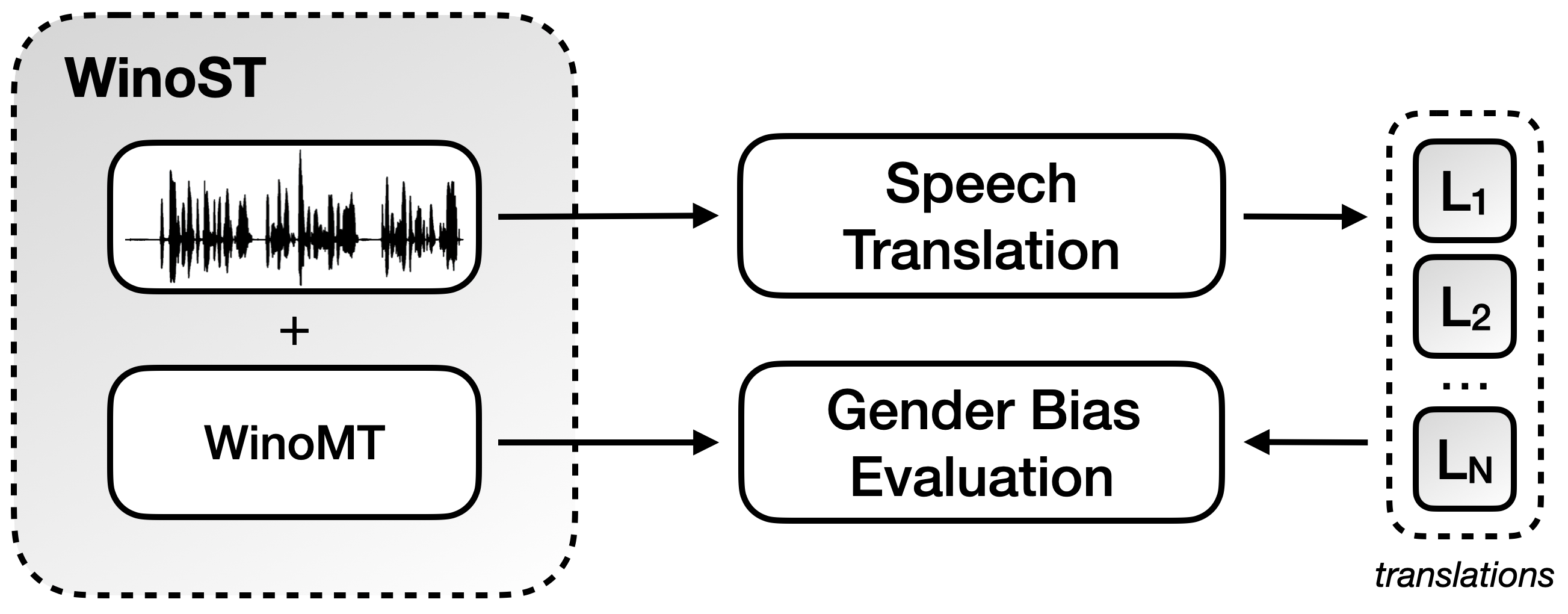}
  \caption{WinoST Evaluation Block Diagram for Speech Translation}
  \label{fig:winost}
\end{figure}

WinoST serves as an input of the ST system to be evaluated, and the output text of the systems follows the same evaluation protocol as WinoMT \cite{gabriel:2019}. Figure \ref{fig:winost} shows the block diagram of this procedure. As a side-product, and not shown in the figure, WinoST can also be used as a challenge set for evaluating ASR gender bias.

Further technical details on WinoST are reported in Table \ref{tab:winost}, including number of files, total hours/words, audio recording and format. The voice mastering process we applied to the recordings includes dynamic voice processing, broadcasting, equalization and filtering. WinoST is available under the MIT License\footnote{\url{https://github.com/gabrielStanovsky/mt\_gender/blob/master/LICENSE}} with the limitation that recordings cannot be used for speech synthesis, text to speech, voice conversion or other applications where the speaker voice is imitated or reproduced.

\begin{table}[th]
  \caption{WinoST details.}
  \label{tab:winost}
  \centering
  \begin{tabular}{ll}
    \toprule
    \textbf{\# Files} & $3,888$ \\
    \midrule
    \textbf{\# Hours} & $\sim6$ \\
    \midrule
    \textbf{\# Words} & $\sim50,500$ \\
    \midrule
    \textbf{Audio format} & WAV ($48$ KHz, $16$-bit) \\
    \bottomrule
  \end{tabular}

\end{table}

\section{Experiments}
\label{sec:evaluation}

In this section, we are describing the first experiments with WinoST. {%\color{red} 
We limit our experiments to four language pairs, but WinoST is extendable to any language pair with English as source language. The only requirement is to have a part-of-speech for the target language.} We describe the baseline ST systems that we are using and the results that we obtain in gender accuracy.

\subsection{Data preprocessing}

Before training the model, we preprocessed both speech and text data. We extracted 40-dimensional log-Mel spectrograms from the audio files, using a window size of 25 ms and hop length of 10 ms, with XNMT \cite{neubig2018xnmt}.\footnote{https://github.com/neulab/xnmt} We normalized the punctuation from text data, we tokenized it, and we de-escaped special characters, using the Moses scripts.\footnote{https://github.com/moses-smt/mosesdecoder} Furthermore, in the case of transcriptions, we lowercased them, and we removed the punctuation. We used the BPE algorithm \cite{sennrich-etal-2016-neural} for encoding translation texts, using a vocabulary size of 8000 for each language, but a character-level encoding in the case of transcriptions.

\subsection{System Details}

The model we used has two convolutional layers with a kernel size of 3, 64 channels and a stride of 2. The Transformer has an embedding size of 512, 6 layers at the encoder and decoder, 8 self-attention heads, and a feed-forward network hidden size of 1024.
We trained the S-Transformer with an Adam optimizer, with a learning rate of $5\cdot10^{-3}$, and an inverse square root scheduler. The training has a warm-up stage of 4000 updates, in which the learning rate grows from $3\cdot10^{-4}$. We used a cross-entropy loss with label smoothing by a factor of 0.1. Moreover, a dropout of 0.1 and a gradient clipping to 20 were applied. Furthermore, we generated the outputs with a beam search of size 5. We loaded 8 sentences per update, with a frequency of 64, which supposes an effective batch size of 512. Those audios longer than 14 seconds and sentences with more than 300 tokens weren't used during training.

\begin{table*}[th]
  \caption{Examples of outputs for different sentences from the WinoST corpus. Words in \textbf{\textcolor{blue}{blue}}, \textbf{\textcolor{red}{red}}, and \textbf{\textcolor{orange}{orange}} indicate male, female and neutral entities, respectively.}
  \label{tab:examples}
  \centering
  \begin{tabular}{r p{0.3\linewidth} p{0.3\linewidth} p{0.29\linewidth}}
    \toprule
     & \textbf{Source transcription} & \textbf{[Target lang.] Predicted translation} & \textbf{Phenomenon} \\
    \midrule
    
    1 & \textbf{\textcolor{orange}{The nurse}} sent the carpenter to the hospital because \textbf{\textcolor{blue}{he}} was extremely cautious & [ES] \textbf{\textcolor{blue}{El enfermero}} envió el carpintero al hospital porque era extremadamente consciente & The gender of “nurse” is correctly inferred from the coreference. \\
    \midrule

    2 & \textbf{\textcolor{orange}{The construction worker}} asked the nurse for an examination because \textbf{\textcolor{red}{she}} was injured & [ES] \textbf{\textcolor{blue}{El trabajador de la construcción}} le pidió a la enfermera una examinación porque estaba \textbf{\textcolor{red}{herida}} & “Construction worker” is stereotyped to its male inflection, with a mismatched adjective gender. \\
    \midrule
    
    3 & The construction worker asked \textbf{\textcolor{orange}{the nurse}} for an examination because \textbf{\textcolor{blue}{he}} was the best & [ES] El trabajador de la construcción le pidió a \textbf{\textcolor{red}{la enfermera}} una examinación porque \textbf{\textcolor{blue}{él}} era \textbf{\textcolor{blue}{el mejor}} & 
    “Nurse” is stereotyped to its female inflection, with a mismatched pronoun and adjective gender. \\
    \midrule
    
    4 & \textbf{\textcolor{orange}{The farmer}} asked the designer what \textbf{\textcolor{red}{she}} could do to help	& [FR] \textbf{\textcolor{blue}{L'agriculteur}} a demandé au designer ce qu'\textbf{\textcolor{red}{elle}} pouvait faire pour aider & Although the pronoun is translated correctly, “farmer” has a biased translation form. \\
    \midrule
    
    5 & The writer wrote a book about \textbf{\textcolor{orange}{the carpenter}} because \textbf{\textcolor{red}{her}} story is very moving & [FR] L'écrivain a écrit un livre sur \textbf{\textcolor{blue}{le charpentier}}, parce que \textbf{\textcolor{orange}{son}} histoire est très émouvante & Biased form for “carpenter” with a neutral possessive gender \\

    \bottomrule
  \end{tabular}
  
\end{table*}

\begin{table}[th]
  \caption{WinoMT Gender Evaluation for four language pairs. Acc.(\% of instances the translation had the correct gender)(the higher the better) \textbf{$\bigtriangleup G$} notes difference in F1 score between masculine and feminine sentences (the higher the worse) and \textbf{$\bigtriangleup S$} notes difference in accuracy between pro/anti stereotypical sentences (the higher the worse).}
  \label{tab:winomt_accuracies}
  \centering
  \begin{tabular}{lrrr}
    \toprule
    \textbf{ST} & \textbf{Acc. ($\uparrow$)} & \bm{$\bigtriangleup G$} ($\downarrow$) & \bm{$\bigtriangleup S$} ($\downarrow$)\\
    \midrule
    \emph{en-de} & $51.0$ & $1.7$ & $1.5$ \\
    \emph{en-es} & $45.2$ & $25.7$ & $12.3$ \\
    \emph{en-fr} & $43.2$ & $13.7$ & $14.5$ \\
    \emph{en-it} & $37.3$ & $23.6$ & $5.6$ \\
    \bottomrule
  \end{tabular}
  
\end{table}

\subsection{Results}

This section describes the results of evaluating the ST system on WinoST and its performance in terms of gender. We are also interested in evaluating ASR English transcriptions and perceive if they contain any gender bias.
\begin{table}[th]
  \caption{WER and BLEU (\%) scores for the MuST-C corpus.}
  \label{tab:asr_accuracies}
  \centering
  \begin{tabular}{lrr}
    \toprule
    \textbf{Language}  & \textbf{ASR (WER $\downarrow$)} & \textbf{ST (BLEU $\uparrow$)} \\
    \midrule
    \emph{en-de} & $24.24$ & $17.8$ \\
    \emph{en-es} & $24.76$ & $21.9$ \\
    \emph{en-fr} & $23.98$ & $28.2$ \\
    \emph{en-it} & $24.18$ & $18.3$ \\
    \bottomrule
  \end{tabular}
  
\end{table}

\paragraph{General ASR and ST Evaluation:}
We use the standard WER and BLEU measures to report the ASR and ST performance, respectively in Table \ref{tab:asr_accuracies}. Our results concur with the results in \cite{mustc19}. 

\paragraph{Gender Bias Evaluation in ST:}
Our main objective is evaluating the accuracy of the systems for each of the language pairs. The high accuracy demonstrates that the system is able to translate the gender of the entities correctly. We also report \bm{$\bigtriangleup G$} and \bm{$\bigtriangleup S$} in Table \ref{tab:winomt_accuracies}. Ideally, these values should be close to 0. High \bm{$\bigtriangleup G$} indicates that the system translates males better, and high \bm{$\bigtriangleup S$} denotes that the system tends to translate pro-stereotypical entities better than anti-stereotypical entities.

The English-to-German (en-de) system has the highest accuracy 51\%. This system also shows the minor difference in treating males and females translations (lowest \bm{$\bigtriangleup G$}, 1.7) and the minor difference in the pro-stereotypical and the anti-stereotypical entities (lowest \bm{$\bigtriangleup S$}, 1.5). The surprising behaviour comes with the English-to-Italian (en-it) system, which has the lowest accuracy of 37.3\%, but still performs reasonably towards the anti-stereotypical entities translations, with the second lowest \bm{$\bigtriangleup S$} difference (5.6). However, the system still favors the male translations with a high \bm{$\bigtriangleup G$} difference (23.6). Both English-to-Spanish (en-es) and English-to-French (en-fr) have similar accuracies (45.2 and 43.2, respectively). However, there is a big difference in the \bm{$\bigtriangleup S$}, which is much higher in the case of en-es (25.7), showing higher bias towards male translations. With these accuracy results, we are showing that the four translation directions present a significant amount of bias and they are far from approaching gender parity in performance. 
Moreover, after manually investigating the translation outputs, we observe that some professions are not correctly translated. \textit{'nurse'} is always translated to the female version in en-es and en-it, and similarly, \textit{'developer'} is always translated to the male version in en-it and en-fr, showing that stereotypes are perpetuated in ST.  Many inflection errors, as illustrated in Table \ref{tab:examples}, 
can occur due to these stereotyped translations. Example 1 shows an anti-stereotypical co-reference case of 'nurse'. One of the common errors happens when translating the gendered adjective or pronoun correctly according to the context while referencing the wrong gendered stereotyped profession, shown in example 2 and 4. Another common problem is mismatched pronouns, the translation of the noun contradicts the profession's translation, due to biased translation in one of them, e.g. example 3. Example 5 shows a biased translation with a neutral pronoun. % Examples 1, 2 are examples of the stereotyped biased translations of the professions neglecting the context.

\paragraph{Gender Bias in ST vs MT:}
Using the S-Transformer, the gender accuracy in the four languages is lower than the reported accuracy of MT commercial systems in the original WinoMT paper \cite{gabriel:2019}. The best reported accuracies from commercial systems reached 74.1\% in en-de, 59.4 \% in en-es, 63.6\% in en-fr and 42.4\% in en-it, while in ST case, it is lower for all language pairs as shown in Table \ref{tab:asr_accuracies}. 

%results are much more biased than in MT commercial systems reported in the original WinoMT paper \cite{gabriel:2019}, where best accuracies from commercial systems reached 74.1\% in en-de, 59.4 \% in en-es, 63.6\% in en-fr and 42.4\% in en-it. 
This may be due to the fact that ST is much more challenging than MT, and lower system performance implies higher biases. This big gap is reduced when comparing in terms of \bm{$\bigtriangleup G$} and \bm{$\bigtriangleup S$}. In this case, ST becomes closer to MT (when comparing in absolute terms), showing even better results in: \bm{$\bigtriangleup G$} for en-it (in MT, 27.8); \bm{$\bigtriangleup S$} for en-de (in MT, 12.5) and en-it (in MT, 9.4).

\paragraph{Gender Bias Evaluation in ASR:}
ASR systems have demonstrated gender biases for female speakers outputs \cite{tatman-2017-gender}.
%ASR systems contain gender biases, e.g., they perform better for male than female speakers \cite{tatman-2017-gender}. 
However, gender bias associated with the context has not been studied in ASR yet, and WinoST allows this analysis. We may expect that ASR is less prone to show gender bias in contextual patterns because of the nature of the task, which inherently combines the purpose of acoustic and language modeling. The acoustic part does not consider long context information, but it tends to benefit from local context information \cite{Sperber2018}. However, the language modeling part takes into consideration the long-range context, and thus it may induce bias \cite{bordia-bowman-2019-identifying,basta_modeling:2021}.

For further analysis of employing WinoST for ASR gender bias evaluation, it is required to distinguish between the gender's errors in transcriptions and the general ones. Therefore,  we computed the global accuracy in WinoST for the ASR best system in Table \ref{tab:asr_accuracies}, en-fr, and got a $74.5\%$ accuracy. However, this global accuracy includes $680$ misspelled professions. Discarding these misspelling errors, we obtained a $98.72\%$ accuracy of predicting pronouns, showing that the amount of gender bias at the context level is quite low in ASR.

%Being interested in the performance of ASR and how it treats the professions of WinoMT.  Thus, arose the need to investigate the bias existing in the ASR transcriptions. Some professions are misspelled since the system was not trained on such professions, so the system tried to find the most probable one that is far from the correct profession. However, we did not count this as bias; we considered it a lack of accuracy. We computed an accuracy score, which has been computed on the number of the correct sentences in the set. The sentence is considered correct if it has the golden entity's profession of the corresponding WinoMT sentence and the correct pronoun representing the correct gender specified for the same entity. The ASR en-fr system's performance, representing the highest accuracy among the four language pairs, reaches 74.15\%, with 680 misspelled professions.
%We found that gender bias occurs when the profession is correct in the sentence, but the pronoun is mispredicted.  Gender accuracy reaches 96.34\% in the same en-fr system.

\section{Conclusions}

This paper presents a new freely available challenge set for evaluating gender bias in ST. This challenge set, WinoST, can benefit from the evaluation protocol, which is widely used for MT. Our set is only based on evaluating the gender inaccuracies in translations in ST systems, mainly relying on the gender information extracted from the context and not from the audio signal.

We used an S-Transformer end-to-end ST system and evaluated their accuracy in terms of gender bias with this new challenge set. Results show that gender accuracy is much lower for ST than for MT, but we have to take into account that ST has also a lower quality than MT. Finally, we show that ASR can exhibit a little gender bias at the contextual-level.

WinoST shares similar limitations as WinoMT, which is the fact of using a synthetic challenge set. Having a synthetic set is positive because of providing a controlled evaluation, and also it is negative because we might be introducing some artificial biases. Therefore, further work could find in the wild transcriptions (with parallel speech utterances) that hold the valuable patterns designed in WinoMT, following \cite{levy_BUG_GB:2021}. 

As mentioned, future work includes making a challenge set from real-life examples, which excerpts human data bias. These examples should be covered by different speakers for studying the effect of gender of such speakers. Following the study of correlation of BLEU and gender bias accuracy \cite{kocmi:2020}, it would be useful to study the correlation between ASR quality and gender bias accuracy. Another approach is to study whether the accent of the speaker imposes bias in the translations. We assume that different accents may have an impact on the ST system. In this paper, we have focused on creating a benchmark for gender bias assessment, we are considering binary gender only. However, there is a need in the community for creating benchmarks for other types of biases, e.g. non-binary gender, racial bias \cite{Koenecke7684}.

\section{Acknowledgements}

This work is supported by the European Research Council (ERC) under the European Union’s Horizon 2020 research and innovation programme (grant agreement No. 947657), the Catalan Agency for Management of University and Research Grants (AGAUR) through the FI PhD Scholarship, Universitat Politècnica de Catalunya with the collaboration of Banco de Santander and the Spanish Ministerio de Ciencia e Innovación through the project PID2019-107579RB-I00 / AEI / 10.13039/501100011033.

\section{References}
\label{lr:ref}
\bibliographystyle{lrec2022-bib}
\bibliography{lrec2022-example}

\bibliographystylelanguageresource{lrec2022-bib}
\bibliographylanguageresource{languageresource}

\end{document}